\documentclass[10pt,twocolumn,letterpaper]{article}

\usepackage[pagenumbers]{cvpr} 
\usepackage{booktabs}
\usepackage{xcolor}
\usepackage{colortbl}
\usepackage{multirow}
\usepackage{pifont}
\usepackage{algorithm}
\usepackage{algpseudocode}
\usepackage[misc]{ifsym}
\newcommand{\cmark}{\ding{51}}
\newcommand{\xmark}{\ding{55}}
\usepackage{stfloats}
\definecolor{cvprblue}{rgb}{0.21,0.49,0.74}
\usepackage[pagebackref,breaklinks,colorlinks,allcolors=cvprblue]{hyperref}

\newcommand\blfootnote[1]{%
  \begingroup
  \renewcommand\thefootnote{}\footnote{#1}%
  \addtocounter{footnote}{-1}%
  \endgroup
}

\title{TempR1: Improving Temporal Understanding of MLLMs via Time-Aware Multi-Task Reinforcement Learning}

\author{Tao~Wu\textsuperscript{1}  \quad \quad Li~Yang\textsuperscript{2} \quad \quad Gen~Zhan\textsuperscript{2}  \quad \quad Yabin~Zhang\textsuperscript{2}\quad \quad Yiting~Liao\textsuperscript{2} \\ Junlin~Li\textsuperscript{2}  \hfill Deliang~Fu\textsuperscript{2}   \hfill Li~Zhang\textsuperscript{2} \hfill Limin~Wang\textsuperscript{1,3,~\Letter}\\
$^1$Nanjing University \quad $^2$ByteDance Inc. \quad $^3$Shanghai AI Lab}

\begin{document}
\maketitle
\begin{abstract}
Enhancing the temporal understanding of Multimodal Large Language Models (MLLMs) is essential for advancing long-form video analysis, enabling tasks such as temporal localization, action detection, and time-sensitive question answering. While reinforcement learning (RL) has recently been explored for improving temporal reasoning, existing approaches are often confined to limited task types and data, restricting their generalization across diverse temporal understanding scenarios. To address this challenge, we present \textbf{TempR1}, a temporal-aware multi-task reinforcement learning framework that systematically strengthens MLLMs’ temporal comprehension. We curate a multi-task corpus that exposes the model to diverse temporal structures and semantics, and build upon the Group Relative Policy Optimization (GRPO) algorithm to achieve stable and effective cross-task optimization. Specifically, we categorize temporal tasks into three correspondence types between predicted intervals and ground-truth instances, and design tailored localization rewards for each, enabling TempR1 to capture fine-grained temporal dependencies and adapt to different temporal patterns. Extensive experiments demonstrate that TempR1 attains state-of-the-art performance across multiple benchmarks. Moreover, its joint optimization over complementary tasks yields a strong synergistic effect, enhancing both generalization and single-task performance, establishing a scalable and principled paradigm for temporal reasoning in MLLMs.
\end{abstract}
\blfootnote{Work done during an internship at ByteDance.}

\section{Introduction}
\label{sec:intro}
\begin{figure}[t]
       \centering
       \includegraphics[width=0.9\linewidth]{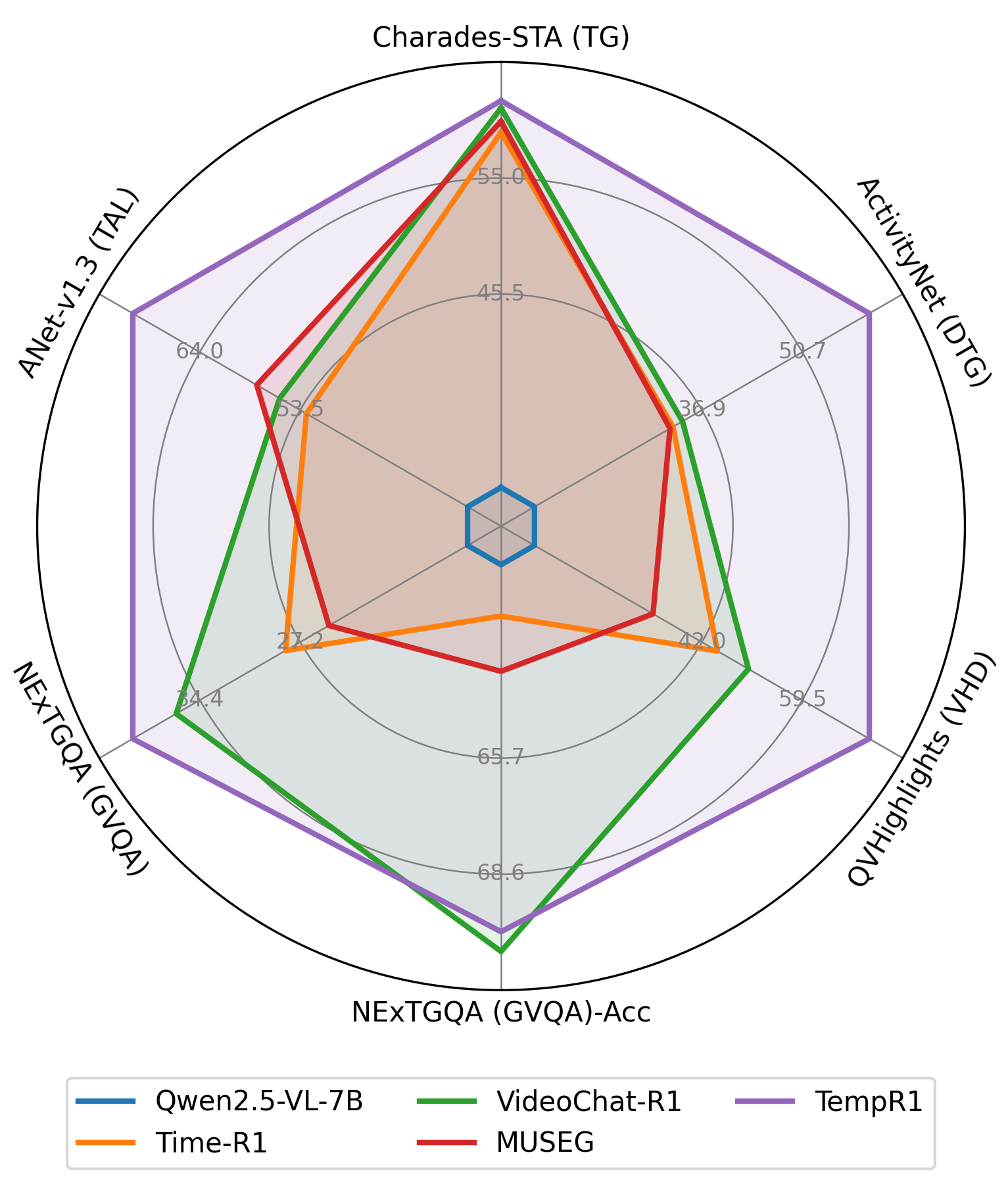}
       \vspace{-8pt}
       \caption{Performance comparison across five temporal understanding tasks.  Our proposed TempR1 achieves the best overall results.}
       \label{fig:radar}
       \vspace{-15pt}
\end{figure}

Understanding temporal dynamics and semantics in videos is fundamental for comprehensive video comprehension. Temporal video understanding encompasses a variety of tasks that require models to reason about when and how events occur and evolve over time., such as Temporal Grounding (TG)~\cite{charadesta,anetgrounding}, Dense Temporal Grounding (DTG)~\cite{anetgrounding,densegrounding}, Temporal Action Localization (TAL)~\cite{thumos,activitynet}, Video Highlight Detection (VHD)~\cite{qvhighlights}, and Grounded Video Question Answering (GVQA)~\cite{nextgqa}. This capability is essential for real-world applications including intelligent video retrieval, human–computer interaction, and long-form video analysis.

Traditional temporal understanding methods typically rely on vision-language pretraining (VLP) models for feature extraction, followed by task-specific prediction modules. While effective on certain benchmarks, such pipelines suffer from feature misalignment and cumulative error propagation between separately trained components. Moreover, their limited cross-task and cross-domain generalization often requires training a distinct model for each dataset or task, hindering scalability and flexibility.

The emergence of MLLMs~\cite{gemini,qwen25vl,gpt4,llava} has shifted research toward end-to-end multimodal reasoning, enabling unified video–text processing within a single architecture. Beyond architectural unification, MLLMs naturally support knowledge transfer across diverse tasks and datasets, opening new opportunities for multi-task and cross-domain temporal understanding without the need for handcrafted task-specific designs. Current MLLM-based approaches for temporal understanding mainly fall into two paradigms: supervised fine-tuning (SFT) and reinforcement learning (RL).

SFT-based methods~\cite{timesuite,trace,vtimellm,timechat} enhance temporal comprehension through large-scale instruction tuning, but they often suffer from overfitting on limited temporal datasets and a loss of general reasoning ability due to rigid supervision~\cite{timer1,videochat-r1}. In contrast, RL-based approaches~\cite{videochat-r1,timer1,tar-tvg,museg} directly optimize task-specific objectives and reward temporally and semantically plausible predictions, leading to better data efficiency and generalization. However, most existing RL methods remain constrained to a narrow scope of temporal understanding, typically temporal grounding, which limits their generalization to broader reasoning scenarios such as dense grounding, action localization, and time-sensitive question answering. Moreover, their lack of exposure to diverse temporal structures prevents them from capturing the hierarchical and compositional nature of temporal dependencies.

To address these limitations, multi-task reinforcement learning provides an effective solution by jointly training on diverse and complementary temporal understanding tasks. These tasks differ in both query semantics and prediction objectives. For instance, TG focuses on event descriptions, TAL on action categories, and VHD on affective or importance-based cues; TG requires single-segment localization, while TAL involves multi-instance prediction and GVQA combines localization with reasoning-based answering. Leveraging such diversity enables the model to adapt to various temporal reasoning demands, enhancing its versatility and generalization. Meanwhile, these tasks share common requirements such as accurate temporal concept modeling, precise timestamp prediction, and coherent video–text alignment. Exploiting these shared structures enriches the learning of fundamental temporal reasoning abilities, resulting in stronger temporal understanding.

In this paper, we propose \textbf{TempR1}, a temporal-aware multi-task reinforcement learning framework that systematically enhances MLLMs’ temporal reasoning across diverse video understanding tasks. We curate a multi-task corpus covering five representative tasks—TG, DTG, TAL, VHD, and GVQA—with over 60K curated samples, exposing the model to rich and diverse temporal event structures. Methodologically, TempR1 builds upon the Group Relative Policy Optimization (GRPO) algorithm to enable efficient and stable cross-task optimization. To account for varying temporal properties across tasks, we categorize them into three types based on the correspondence between predicted intervals and ground-truth instances and design tailored reward functions for each. This structured reward design allows TempR1 to capture fine-grained temporal relationships and adapt effectively to different temporal structures. As shown in Figure~\ref{fig:radar}, extensive experiments across multiple benchmarks demonstrate that TempR1 achieves state-of-the-art performance on the five temporal understanding tasks. Notably, the joint reinforcement optimization not only improves multi-task generalization but also enhances performance on individual tasks, highlighting the synergistic effect of our unified design. Our work establishes a principled and scalable paradigm for multi-task reinforcement learning in video temporal understanding, paving the way toward MLLMs with stronger and more generalizable temporal reasoning capabilities.
\section{Related Work}
\label{sec:relatedwork}

\noindent \textbf{Video Temporal Understanding.} 
Video temporal understanding~\cite{charadesta,anetgrounding,activitynet,densegrounding,nextgqa,qvhighlights} aims to analyze the temporal structure and semantic dynamics of videos. Early approaches~\cite{umt,qd-detr,univtg,eatr,momentdiff,tr-detr,cg-detr,uvcom,r2tuning,flashvtg,prvg,hscnet} primarily rely on VLP models~\cite{clip,videoclip,internvideo,internvideo2,unmasked} to extract vision and text representations, followed by cross-modal attention and task-specific heads for objective prediction. While effective, these task-specific architectures require distinct expert models for each dataset, limiting scalability and cross-task generalization.

Recent advances in MLLMs~\cite{gemini,qwen25vl,gpt4,llava,videochat,mvbench,videochat-online,llava-onevision,video-llava,video-llama} have demonstrated strong open-vocabulary understanding and multimodal reasoning capabilities. However, due to limitations in pretraining objectives and data, they still struggle to capture fine-grained temporal dependencies and align linguistic semantics with temporal visual cues. Supervised fine-tuning (SFT) has been explored to enhance temporal comprehension~\cite{chatvtg,wang2024efficient,groundinggpt,trace,moviechat,timesuite,vtimellm,timechat,hawkeye,etbench}. Yet, its rigid token-level supervision tends to cause overfitting to instruction data.

In contrast, RL-based fine-tuning directly optimizes task-specific objectives and rewards all reasonable predictions, improving data efficiency and generalization. Recent RL approaches~\cite{videochat-r1,timer1,museg,tar-tvg,videochat-r15} achieve strong performance on individual temporal understanding tasks, but mostly remain limited to temporal grounding and lack a unified framework for broader temporal tasks. In this work, we extend RL-based temporal understanding to multiple tasks and propose a scalable reward formulation that adapts to diverse temporal correspondence patterns. By jointly training across these tasks, our proposed \textbf{TempR1} achieves consistent improvements and establishes a generalizable paradigm for enhancing MLLMs’ temporal understanding.

\noindent \textbf{MLLM Reinforcement Fine-tuning.} 
Following the success of RL in enhancing the reasoning abilities of LLMs through verifiable reward mechanisms~\cite{openaio1,deepseekr1}, recent works have extended RL optimization to MLLMs. These methods~\cite{shen2025vlm,liu2025visual,chen2025visrl,xia2025visionary,huang2025vision,wang2025videorft,feng2025video,bai2025univg,tang2025tspo,zhang2025tinyllava,ouyang2025spacer,chen2025scaling,sc-captioner} design task-specific, rule-based rewards to strengthen multimodal reasoning, leading to improvements in perception, visual grounding, and compositional understanding. However, most of them focus on static images or short video clips, with limited exploration of long-form temporal reasoning tasks beyond temporal grounding.

\begin{figure*}[t]
       \centering
       \includegraphics[width = 0.94\textwidth]{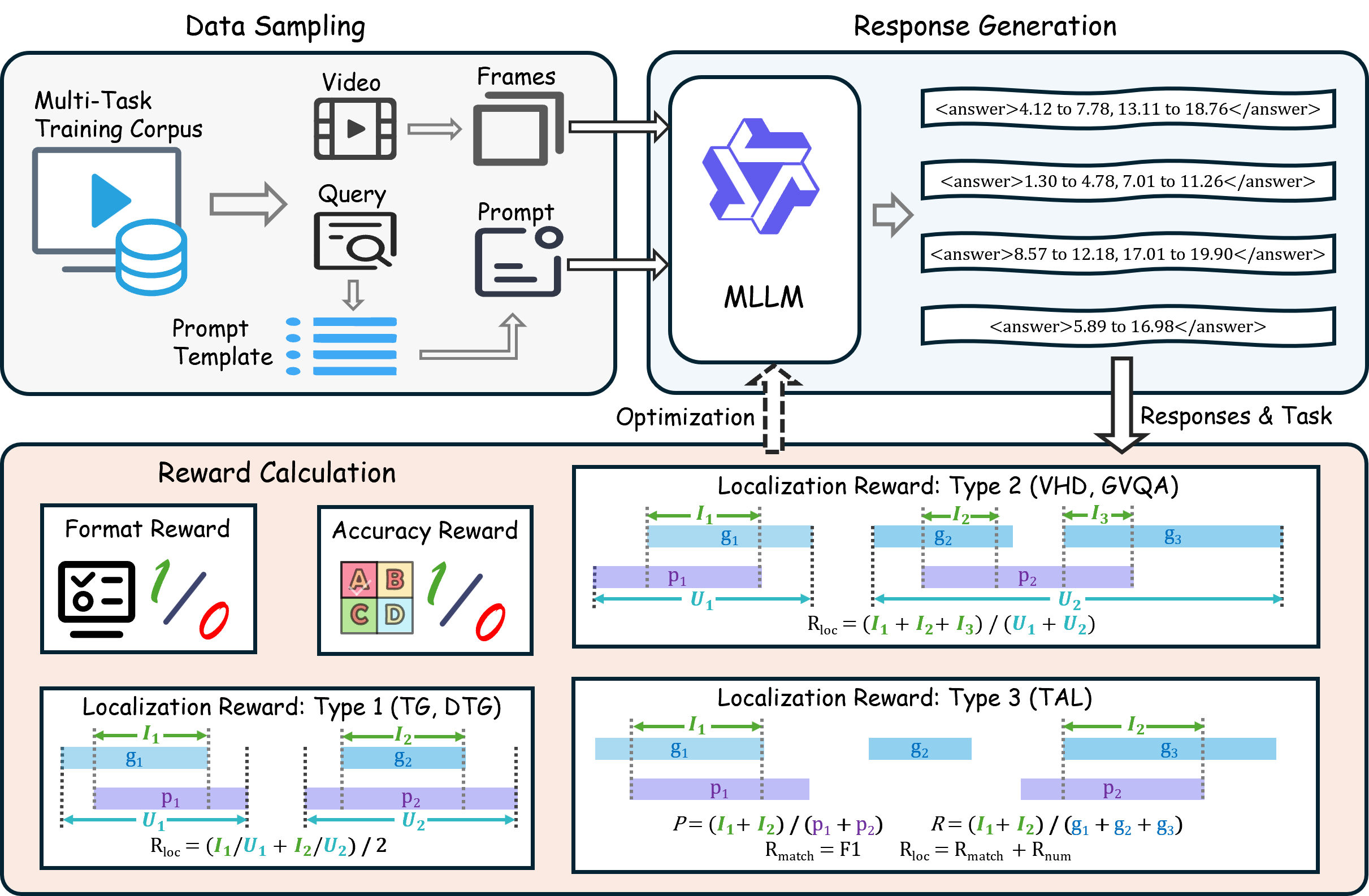}
       \caption{Overview of the \textbf{TempR1} framework. We finetune the MLLM on a multi-task training corpus covering five temporal understanding tasks. Reinforcement learning is applied with rule-based rewards, including format and accuracy rewards, as well as localization rewards for three correspondence types: Type~1 (one-to-one, TG/DTG), Type~2 (many-to-one, VHD/GVQA), and Type~3 (many-to-many, TAL). These rewards jointly optimize the MLLM to achieve accurate and robust temporal prediction.}
       \label{fig:overview}
\end{figure*}

\section{Method}

\subsection{Revisiting Group Relative Policy Optimization}
\label{subsec:grpo}

Group Relative Policy Optimization (GRPO)~\cite{deepseekr1} extends Proximal Policy Optimization (PPO) by replacing critic-based value estimation with \textit{group-wise relative comparison}, eliminating the need for a separate critic network and greatly reducing training overhead. Specifically, given an input prompt $p$, the policy model $\pi_\theta$ samples a group of $G$ candidate responses $o = \{o_1, \dots, o_G\}$. Each response is evaluated by a reward function $r(\cdot)$, yielding the corresponding scores $\{ r(o_1), \dots, r(o_G) \}$. Instead of learning an absolute value function, GRPO normalizes these scores within the group to obtain \textit{relative advantages} that reflect the comparative quality of each response:
\begin{equation}
\label{eq:grpo-adv}
A_i = 
\frac{r(o_i) - \mathrm{mean}\big(\{r(o_j)\}_{j=1}^G\big)}
     {\mathrm{std}\big(\{r(o_j)\}_{j=1}^G\big)} ,
\end{equation}
where a higher $A_i$ indicates a better response within the sampled group.  The policy is then optimized to increase the likelihood of higher-quality responses. The GRPO objective is formulated as:
\begin{equation}
\label{eq:grpo-obj}
\begin{aligned}
\max_{\pi_\theta} \, \mathbb{E}_{o \sim \pi_{\theta_{\text{old}}}} \Big[
&\sum_{i=1}^G \min \big( r_i \cdot A_i, \, \text{clip}(r_i, 1-\epsilon, 1+\epsilon) \cdot A_i \big) \\
&- \, \beta \, \mathrm{D}_{\mathrm{KL}}\big(\pi_\theta \parallel \pi_\mathrm{ref}\big)
\Big],
\end{aligned}
\end{equation}
where $r_i = \tfrac{\pi_\theta(o_i)}{\pi_{\theta_{\text{old}}}(o_i)}$, and the KL-divergence term regularizes the updated policy to remain close to a reference policy $\pi_\mathrm{ref}$, ensuring stable optimization.

\subsection{Task Formulation and Format Reward}
\label{subsec:format_reward}

Unlike prior RL-based methods that focus on temporal grounding, TempR1 jointly trains on multiple complementary temporal understanding tasks, leveraging shared temporal reasoning while capturing task-specific temporal and semantic patterns. We consider five tasks:

\noindent \textbf{Temporal Grounding (TG)} locates the interval of an event described by a query: 
$O=\textless\mathrm{answer}\textgreater t_s\ \mathrm{to}\ t_e\textless\mathrm{/answer}\textgreater$.

\noindent \textbf{Dense Temporal Grounding (DTG)} localizes multiple intervals corresponding to sequential events: 
$O=\textless\mathrm{answer}\textgreater t_{s1}\ \mathrm{to}\ t_{e1},\, t_{s2}\ \mathrm{to}\ t_{e2},\dots\textless\mathrm{/answer}\textgreater$.

\noindent \textbf{Video Highlight Detection (VHD)} identifies an uncertain number of highlight segments describing the same event, sharing the same output format as DTG.

\noindent \textbf{Grounded Video QA (GVQA)} predicts both the answer and supporting evidence: 
$O=\textless\mathrm{answer}\textgreater A\textless\mathrm{/answer}\textgreater
\textless\mathrm{glue}\textgreater t_s^1\ \mathrm{to}\ t_e^1,\dots\textless\mathrm{/glue}\textgreater$.

\noindent \textbf{Temporal Action Localization (TAL)} localizes all instances of a predefined action, also using the DTG format.

To ensure consistent supervision, we define a \textbf{format reward} $R_{\text{format}}$ based on regular-expression matching between the model output and the required template:
\begin{equation}
R_{\text{format}}(o)=
\begin{cases}
1, & \text{if } o \text{ matches the format},\\
0, & \text{otherwise.}
\end{cases}
\end{equation}
This encourages the model to produce temporally grounded, machine-parsable outputs for stable multi-task reinforcement learning.

\subsection{Temporal Localization Reward}
\label{subsec:localization_reward}

As shown in Figure~\ref{fig:overview}, to guide accurate temporal localization, we design a set of task-specific \textbf{localization rewards} that measure the alignment between predicted and ground-truth temporal intervals.  
Depending on the correspondence structure between predicted temporal intervals and ground-truth instances, we categorize tasks into three types: one-to-one, many-to-one, and many-to-many.

\noindent \textbf{Type~1: One-to-One Correspondence (TG, DTG).}
Each predicted interval corresponds to a single ground-truth event, and the number of predicted and ground-truth intervals is the same.  
The reward is the mean temporal Intersection-over-Union (IoU) across all corresponding pairs:
\begin{equation}
R_{\text{loc}}^{\text{(TG/DTG)}} =
\frac{1}{N} \sum_{i=1}^{N}
\frac{\mathrm{Intersection}(p_i, g_i)}{\mathrm{Union}(p_i, g_i)} .
\end{equation}

\noindent \textbf{Type~2: Many-to-One Correspondence (VHD, GVQA).}
In the VHD and GVQA tasks, multiple predicted temporal intervals correspond to a single ground-truth instance: in VHD, several highlight segments jointly represent one highlight event, while in GVQA, multiple evidence segments jointly support a single video reasoning question. For such tasks, we treat the predicted and ground-truth temporal segments as unified entities and compute the temporal IoU between these two aggregated intervals.Specifically, we first merge all predicted intervals into a single union region and compute the IoU with the union of ground-truth intervals:
\begin{equation}
R_{\text{loc}}^{\text{(VHD/GVQA)}} =
\frac{\mathrm{Intersection}(\cup_i p_i, \cup_j g_j)}{\mathrm{Union}(\cup_i p_i, \cup_j g_j)} .
\end{equation}

\noindent \textbf{Type~3: Many-to-Many Correspondence (TAL).}
In the TAL task, the model is required to predict all action instances that belong to the queried category. It outputs multiple temporal intervals, each representing one predicted instance. The number of predicted intervals may differ from the number of ground-truth instances, as the model has no prior knowledge of the true instance count. To address this, we design two complementary components within the temporal localization reward to jointly evaluate the accuracy of instance prediction and temporal boundary localization:  

\noindent (1) an \textbf{instance number reward} to penalize mismatches in the number of predictions and groundtruth instances:
\begin{equation}
R_{\text{num}} = \exp\left(-\frac{|N_{\text{pred}} - N_{\text{gt}}|}{\min(N_{\text{gt}}, 3) \cdot \sigma}\right),
\end{equation}
where \( N_{\text{pred}} \) and \( N_{\text{gt}} \) denote the predicted and ground-truth instance counts, respectively. The term \( \min(N_{\text{gt}}, 3) \) ensures robustness across different instance numbers, and \( \sigma \) is a hyperparameter that adjusts the penalty strength for count mismatches. This reward exponentially decreases as the absolute difference between predicted and ground-truth counts increases, yielding a score between 0 and 1 to quantify the accuracy of instance number prediction.

\begin{algorithm}[t]
\caption{Matching between Predicted Intervals and Groundtruth Instances}
\label{alg:dp_f1}
\begin{algorithmic}[1]
\Require Predicted intervals $\{p_i\}_{i=1}^m$, ground truths $\{g_j\}_{j=1}^n$
\State Compute IoU matrix $\mathbf{I}\!\in\!\mathbb{R}^{m\times n}$
\State Initialize DP table $\mathbf{D}\!\in\!\mathbb{R}^{(m+1)\times(n+1)}\!=\!0$, path table $\mathbf{P}$
\For{$i=1\to m$} \For{$j=1\to n$}
    \State $a\!=\!\mathbf{D}[i\!-\!1,j],\ b\!=\!\mathbf{D}[i,j\!-\!1],\ c\!=\!\mathbf{D}[i\!-\!1,j\!-\!1]\!+\!\mathbf{I}[i,j]$
    \State $\mathbf{D}[i,j]\!\leftarrow\!\max(a,b,c)$
    \State Update $\mathbf{P}[i,j]$ according to selected option
\EndFor \EndFor
\State \Return matched pairs $\mathcal{M}=\mathbf{P}[m,n]$
\end{algorithmic}
\end{algorithm}

\noindent (2) a \textbf{matching reward} to evaluate the accuracy of temporal boundary prediction. We first match the predicted temporal intervals with the ground-truth instances. 
The predicted and ground-truth intervals are each sorted in chronological order. 
We assume that earlier predictions should correspond to earlier ground-truth instances, 
which allows us to perform matching using a dynamic programming algorithm as described in Algorithm~\ref {alg:dp_f1}. 
The objective of dynamic programming is to maximize the total IoU between all matched pairs, 
defined as the summed IoU ($s\text{IoU}$):

\begin{equation}
s\text{IoU} = \sum_{(p_i, g_j) \in \mathcal{M}} 
\frac{\mathrm{Intersection}(p_i, g_j)}{\mathrm{Union}(p_i, g_j)},
\end{equation}
where $\mathcal{M}$ denotes the set of matched prediction–ground-truth pairs obtained from dynamic programming. 
Then we compute the precision $P$, recall $R$, and F1 score as:
\begin{equation}
P = \frac{s\text{IoU}}{\text{num}_{\text{pred}}}, \quad
R = \frac{s\text{IoU}}{\text{num}_{\text{gt}}}, \quad
\text{F1} = \frac{2PR}{P + R}.
\end{equation}
The F1 score is used as the matching reward: $R_{\text{match}}=\text{F1}$
The final localization reward for TAL is obtained by combining both components:
\begin{equation}
R_{\text{loc}}^{\text{(TAL)}} =  R_{\text{num}} + R_{\text{match}}.
\end{equation}

\begin{table*}[t]
\resizebox{\textwidth}{!}{
\LARGE
\begin{tabular}{@{}ll|c|cccc|cccc|cccc@{}}
\toprule
\multirow{2}{*}{Type} & \multirow{2}{*}{Method} & \multirow{2}{*}{ZS} & \multicolumn{4}{c|}{Charades-STA (TG)} & \multicolumn{4}{c|}{ActivityNet (TG)} & \multicolumn{4}{c}{QVHighlights (VHD)} \\
 &  &  & mIoU & R1@0.3 & R1@0.5 & R1@0.7 & mIoU & R1@0.3 & R1@0.5 & R1@0.7 & mIoU & R1@0.3 & R1@0.5 & R1@0.7 \\ \midrule
\multirow{8}{*}{VLP} & M-DETR~\cite{qvhighlights} & \xmark & - & - & 55.7 & 34.2 & - 
 & - & - & - & - & - & 53.9 & 34.8 \\
 & UMT~\cite{umt} & \xmark & - & - & 49.4 & 26.2 & - & - & - & - & - & - & 60.3 & 44.3 \\
 & QD-DETR~\cite{qd-detr} & \xmark & - & - & 57.3 & 32.6 & - & - & - & - & - & - & 62.7 & 46.7 \\
 & UniVTG~\cite{univtg} & \xmark & 52.2 & 72.6 & 60.2 & 38.6 & - & 56.1 & 43.4 & 24.3 & - & - & 59.7 & - \\
 & SSRN~\cite{ssrn} & \xmark & - & - & 65.6 & 42.7 & - & - & 54.5 & 33.2 & - & - & - & - \\
 & EaTR~\cite{eatr} & \xmark & - & - & 68.5 & 44.9 & - & - & 58.2 & 37.6 & - & - & 61.4 & 45.8 \\
 & CG-DETR~\cite{cg-detr} & \xmark & 50.1 & 70.4 & 58.4 & 36.3 & - & - & - & - & - & - & 67.4 & 52.1 \\
 & FlashVTG~\cite{flashvtg} & \xmark & - & - & 70.3 & 49.9 & - & - & - & - & - & - & 73.1 & 57.3 \\ \midrule
\multirow{8}{*}{SFT} & ChatVTG~\cite{chatvtg} & \cmark & 34.9 & 52.7 & 33.0 & 15.9 & 27.2 & 40.7 & 22.5 & 9.4 & - & - & - & - \\
 & TimeChat~\cite{timechat} & \cmark & - & - & 32.2 & 13.4 & - & - & - & - & - & - & - & - \\
 & HawkEye~\cite{hawkeye} & \cmark & 33.7 & 50.6 & 31.4 & 14.5 & 32.7 & 49.1 & 29.3 & 10.7 & - & - & - & - \\
 & HawkEye*~\cite{hawkeye} & \xmark & 49.3 & 72.5 & 58.3 & 28.8 & 39.1 & 55.9 & 34.7 & 17.9 & - & - & - & - \\
 & VTimeLLM~\cite{vtimellm} & \cmark & 31.2 & 51.0 & 27.5 & 11.4 & 30.4 & 44.0 & 27.8 & 14.3 & - & - & - & - \\
 & TimeSuite~\cite{timesuite} & \cmark & - & 69.9 & 48.7 & 24.0 & - & - & - & - & - & - & - & - \\
 & TimeSuite*~\cite{timesuite} & \xmark & - & 79.4 & 67.1 & 43.0 & - & - & - & - & - & - & - & - \\
 & TRACE~\cite{trace} &\cmark &-  & - & 40.3 & 19.4 & 39.0 & - & 37.7 & 24.0 & - & - & - & - \\ \midrule
 & Qwen-2.5-VL-7B~\cite{qwen25vl} & \cmark & 29.6 & 46.1 & 25.5 & 11.5 & 17.2 & 22.5 & 13.0 & 6.3 & 12.9 & 11.8 & 6.3 & 3.0 \\ \midrule

\multirow{7}{*}{RL} & Time-R1*~\cite{timer1} & \xmark & 58.8 & 82.8 & 72.2 & 50.1 & - & 73.3 & 55.6 & 34.0 & 44.7 & 63.2 & 47.1 & 28.0 \\
 & TAR-TVG*~\cite{tar-tvg} & \xmark & 61.1 & \textbf{83.6} & 71.4 & 50.2 & - & - & - & - & 65.9 & 85.6 & 76.1 & 58.5 \\
 & TAR-TVG~\cite{tar-tvg} & \xmark & - & - & - & - & \underline{41.1} & \underline{61.5} & \underline{39.8} & \underline{19.8} & - & - & - & - \\
 & VideoChat-R1~\cite{videochat-r1} & \xmark & 60.8 & - & 71.7 & 50.2 & 36.6 & - & 33.4 & 17.7 & 50.1 & 70.5 & 53.7 & 34.1 \\
 & VideoChat-R1.5~\cite{videochat-r15} & \xmark & 60.6 & 82.8 & 71.6 & 48.3 & 20.2 & 28.3 & 15.6 & 7.2 & 18.0 & 24.5 & 11.7 & 4.3 \\
 & MUSEG~\cite{museg} & \xmark & 59.7 & - & - & - & 26.5 & 38.3 & 22.4 & 10.5 & 33.5 & 49.4 & 30.8 & 15.5 \\
 & TempR1  & \xmark & \underline{61.4} & \underline{83.2} & \underline{72.7} & \underline{51.2} & 32.2 & 47.6 &  28.9 & 14.0 & \underline{71.1} & \underline{91.1} & \underline{79.4} & \underline{61.4} \\
 & TempR1* & \xmark & \textbf{62.5} & 83.0 & \textbf{73.2} & \textbf{52.3} & \textbf{51.9} & \textbf{72.5} & \textbf{55.4} & \textbf{34.1} & \textbf{71.6} & \textbf{91.5} & \textbf{80.1} & \textbf{62.0} \\ \bottomrule
\end{tabular}}
\caption{Results on \textbf{Charades-STA} and \textbf{ActivityNet-Caption} for Temporal Grounding (TG), and on \textbf{QVHighlights} for Video Highlight Detection (VHD). TempR1 is compared with both VLP-based expert models and open-source MLLMs. A \cmark\ in the ``ZS" column denotes zero-shot inference without using any data from the three datasets, while * indicates models further fine-tuned on each single dataset.}
\label{tab:sota1}
\end{table*}

\subsection{Multi-task Training with GRPO}
\label{subsec:multi_grpo}

We integrate all rewards into a unified multi-task reinforcement learning framework based on GRPO and select the corresponding combination of Format Reward and Localization Reward based on the task type.
For the GVQA task, an additional classification reward is used to evaluate the accuracy of the answer to the question:
\begin{equation}
R_{\text{cls}} =
\begin{cases}
1, & \text{if the answer option is correct},\\
0, & \text{otherwise.}
\end{cases}
\end{equation}

Formally, for each training sample, given the task type $t$, the overall reward is defined as:
\begin{equation}
R = R_{\mathrm{format}} + R_{\mathrm{loc}}^{(t)} + \mathbf{1}_{\{t=\mathrm{GVQA}\}}\, R_{\mathrm{cls}}.
\end{equation}

\section{Experiments}

\subsection{Training Data}
\label{subsec:training_data}

We build a large-scale multi-task corpus for temporal video understanding, containing about \textbf{60K} video samples collected from multiple datasets. Specifically, TG data come from Charades-STA~\cite{charadesta}, DiDeMo~\cite{didemo}, and TimeRFT~\cite{timer1}; DTG from ActivityNet-Caption~\cite{anetgrounding}; VHD from QVHighlights~\cite{qvhighlights}; GVQA from NExTGQA~\cite{nextgqa}; and TAL from ActivityNet-v1.3~\cite{activitynet} and HACS~\cite{hacs}.

\begin{table*}[t]
\large
\resizebox{\textwidth}{!}{
\begin{tabular}{@{}ll|cccc|ccccc|c@{}}
\toprule
\multirow{2}{*}{Type} & \multirow{2}{*}{Method} & \multicolumn{4}{c|}{ActivityNet (DTG)} & \multicolumn{5}{c|}{NExTGQA (GVQA)} & \multicolumn{1}{c}{ActivityNet-v1.3 (TAL)} \\
 & \multicolumn{1}{c|}{} & \multicolumn{1}{c}{mIoU} & \multicolumn{1}{c}{R@0.3} & \multicolumn{1}{c}{R@0.5} & \multicolumn{1}{c|}{R@0.7} & \multicolumn{1}{c}{mIoU} & R@0.3 & R@0.5 & R@0.7 & \multicolumn{1}{c|}{Acc} & \multicolumn{1}{c}{mF1} \\ \midrule
\multirow{2}{*}{VLP} & PRVG~\cite{prvg} & 55.6 & 78.3 & 61.2 & 37.8 & - & - & - & - & - & - \\
 & HSCNet~\cite{hscnet} & \underline{59.7} & \underline{81.9} & \underline{66.6} & \textbf{44.0} & - & - & - & - & - & - \\ \midrule
\multirow{5}{*}{MLLM} & Qwen2.5-VL-7B~\cite{qwen25vl} & 14.1 & 18.3 & 10.3 & 5.0 & 15.3 & 20.6 & 11.1 & 5.2 & 60.8 & 35.9 \\
& Time-R1~\cite{timer1} & 33.0 & 48.5 & 29.7 & 13.9 & 28.3 & \underline{39.2} & \underline{18.6} & \underline{8.2} & 62.1 & 52.8 \\
 & VideoChat-R1~\cite{videochat-r1} & 34.3 & 50.7 & 31.3 & 15.3 & \underline{36.1} & - & - & - & \textbf{70.6} & 55.6 \\
 & VideoChat-R1.5~\cite{videochat-r15} & 14.6 & 20.0 & 10.9 & 5.2 & 20.5 & 28.2 & 12.7 & 4.6 & 62.4 & 26.8  \\
 & MUSEG~\cite{museg} & 32.6 & 48.5 & 29.5 & 13.5 & 25.2 & 34.7 & 16.4 & 7.4 & 63.5 & 58.0 \\
 & TempR1 & \textbf{59.8} & \textbf{82.6} & \textbf{66.7} & \underline{43.4} & \textbf{39.2} & \textbf{58.9} & \textbf{35.9} & \textbf{17.1} & \underline{70.1} & \textbf{71.0} \\ \bottomrule
\end{tabular}}
\caption{
Results on Dense Temporal Grounding (DTG), Grounded Video Question Answering (GVQA), and Temporal Action Localization (TAL). 
Our TempR1 consistently outperforms prior VLP models and RL-based MLLM methods.
}
\label{tab:sota2}
\end{table*}
\begin{figure}[t]
       \centering
       \vspace{-10pt}\includegraphics[width=\linewidth]{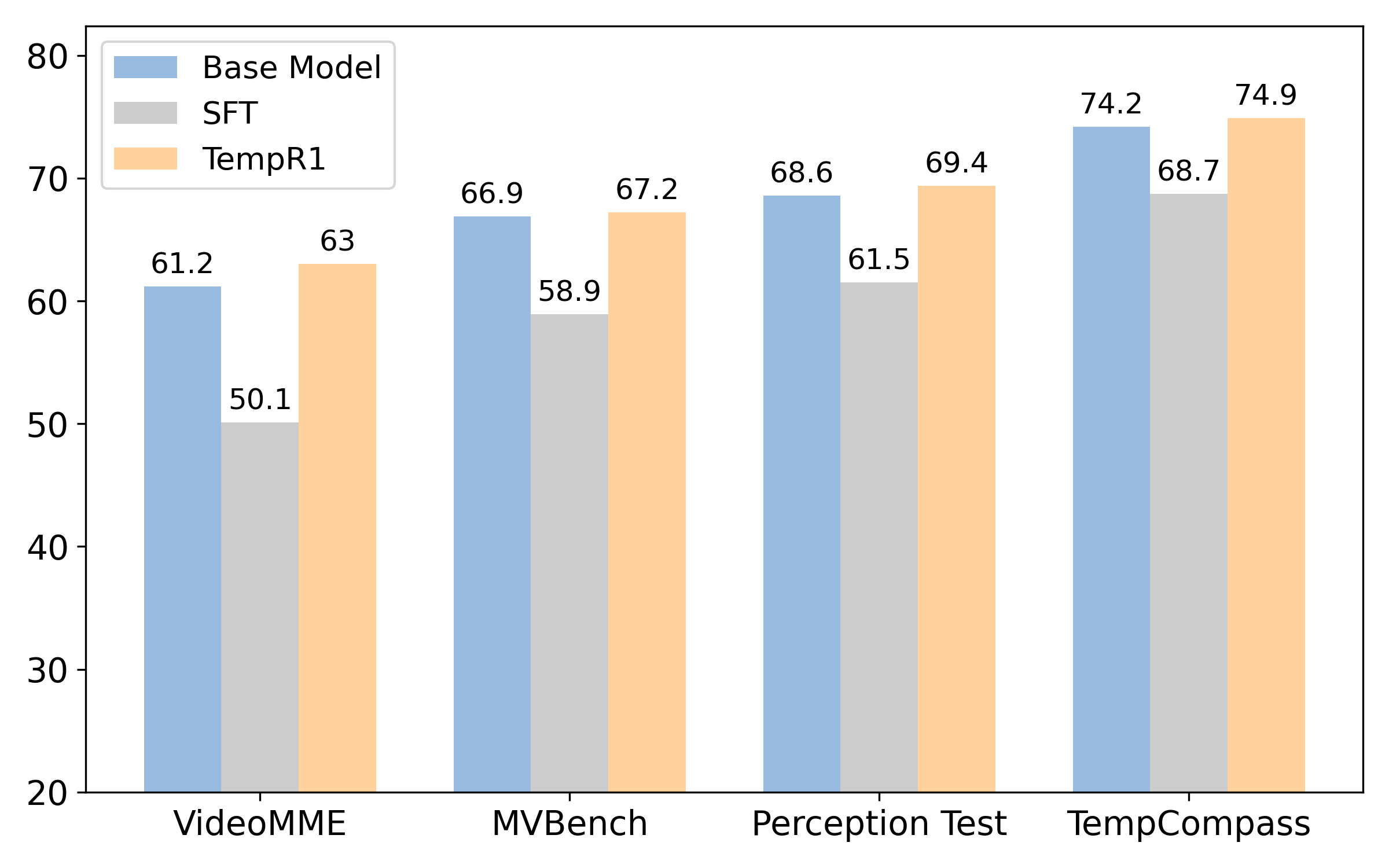}
       \vspace{-20pt}
       \caption{Comparison with the Qwen2.5-VL-7B base model and supervised fine-tuning (SFT) results on general video benchmarks.}
       \label{fig:general}
       \vspace{-10pt}
\end{figure}
\subsection{Benchmarks and Evaluation Metrics}
\label{subsec:benchmarks}
We evaluate TempR1 on five temporal understanding tasks using standard public benchmarks. For \textbf{TG}, evaluations are conducted on Charades-STA~\cite{charadesta} and ActivityNet-Caption~\cite{anetgrounding} using mean IoU (mIoU) and Recall at various IoU thresholds. \textbf{DTG} is evaluated on ActivityNet-Caption~\cite{anetgrounding} with the same metrics. \textbf{VHD} is assessed on QVHighlights~\cite{qvhighlights} following TAR-TVG~\cite{tar-tvg}, where multiple highlight segments are aggregated for IoU computation, and both mIoU and Recall are reported. \textbf{GVQA} is evaluated on NExT-GQA~\cite{nextgqa}, reporting answer accuracy and the mean IoU of supporting evidence. \textbf{TAL} follows MUSEG~\cite{museg} and uses the mean F1 score (mF1) under four IoU thresholds (0.1, 0.3, 0.5, 0.7). To further assess general understanding, we also test on VideoMME~\cite{videomme}, Perception Test~\cite{perceptiontest}, TempCompass~\cite{tempcompass}, and MVBench~\cite{mvbench}.

\begin{table}[t]
\centering
\Large
\vspace{-6pt}
\resizebox{\linewidth}{!}{
\begin{tabular}{cc|ccccc}
\toprule
\multirow{2}{*}{Num Reward} &  \multirow{2}{*}{Match Reward} & \multicolumn{5}{c}{ANet-v1.3 (TAL)} \\
 &  & F1@0.1 & F1@0.3 & F1@0.5 & F1@0.7 & mF1 \\ \midrule
\cmark & Sequential & 84.1 & 53.5 & 29.8 & 14.3 & 45.4 \\
\xmark & DP-based & 89.2 & 76.4 & 63.6 & 50.2 & 69.8 \\
\rowcolor[HTML]{EFEFEF}
\cmark & DP-based & 89.8 & 77.3 & 64.5 & 50.9 & 70.6 \\ \bottomrule
\end{tabular}}
\caption{Ablation study on the design of Type-3 temporal localization reward for the TAL task on ActivityNet-v1.3.}
\vspace{-10pt}
\label{tab:abreward}
\end{table}
\subsection{Implementation Details}

We adopt Qwen2.5-VL-7B~\cite{qwen25vl} as the base model for reinforcement fine-tuning. Following existing works~\cite{museg}, video frames are sampled at 2 FPS, and when the total number of frames exceeds 448, we uniformly sample 448 frames. The spatial resolution of the sampled frames is dynamically scaled so that the total number of visual tokens does not exceed 3,584. During training, we mix data from all five temporal understanding tasks and randomly sample the data batch at each training step. For samples within a batch, task-specific prompts and reward functions are selected according to their task type. The model is trained for one epoch on the 60K multi-task training corpus. During inference, we apply the same frame sampling strategy and task-specific textual prompts as in training. The model outputs are parsed using regular expression matching to extract temporal intervals and structured results.

\begin{table*}[t]
\large
\resizebox{\textwidth}{!}{
\begin{tabular}{l|ccc|ccc|cccc|ccc|c}
\toprule
\multirow{2}{*}{RL Algrithm} & \multicolumn{3}{c|}{Charades (TG)} & \multicolumn{3}{c|}{ActivityNet (DTG)} & \multicolumn{4}{c|}{NExTGQA (GVQA)} & \multicolumn{3}{c|}{QVHighlights (VHD)} & ANet-v1.3 (TAL) \\
 & mIoU & R@0.5 & R@0.7 & mIoU &  R@0.5 & R@0.7 & mIoU & R@0.5 & R@0.7 & Acc & mIoU & R@0.5 & R@0.7 & mF1 \\ \midrule
DAPO~\cite{dapo} & 60.0 & 72.0 & 48.8 & 60.3 & 67.7 & 43.8 & 36.9 & 33.7 & 15.7 & 69.4 & 69.5 & 76.2 & 58.3  & 47.6 \\
GSPO~\cite{gspo} & 62.5 & 72.3 & 50.8 & 59.9 & 66.6 & 43.5 & 38.4 & 34.8 & 17.0 & 70.7 & 72.6 & 80.6 & 63.4 & 71.7 \\
\rowcolor[HTML]{EFEFEF}
GRPO~\cite{deepseekmath} & 61.4 & 72.7 & 51.2 & 59.8 & 66.8 & 43.4 & 39.2 & 35.9 & 17.1 & 70.1 & 71.1 & 79.4 & 61.4 & 71.0 \\ \bottomrule
\end{tabular}}
\caption{Ablation study of different reinforcement learning algorithms for temporal understanding. We compare DAPO, GSPO, and GRPO under the same multi-task training setting. Results show that GSPO and GRPO achieve better performance than DAPO on most tasks.}
\label{tab:abrl}
\end{table*}
\begin{table*}[t]
\Large
\resizebox{\textwidth}{!}{
\begin{tabular}{l|ccc|ccc|cccc|ccc|c}
\toprule
\multirow{2}{*}{Task} & \multicolumn{3}{c|}{Charades (TG)} & \multicolumn{3}{c|}{ActivityNet (DTG)} & \multicolumn{4}{c|}{NExTGQA (GVQA)} & \multicolumn{3}{c|}{QVHighlights (VHD)} & ANet-v1.3 (TAL) \\
 & mIoU & R@0.5 & R@0.7 & mIoU  & R@0.5 & R@0.7 & mIoU & R@0.5 & R@0.7 & Acc & mIoU & R@0.5 & R@0.7 & mF1 \\ \midrule
TG & 60.2 & 71.3 & 48.2 & 16.4 & 10.7 & 4.6 & 21.0 & 14.4 & 4.7 & 62.5 & 4.9 & 0.0 & 0.0 & 19.9 \\
TG+GVQA & 60.1 & 72.0 & 48.9 & 17.7 & 11.9 & 4.9 & 34.9 & 31.9 & 13.6 & 67.5 & 7.1  & 1.6 & 0.0 & 34.3  \\
TG+DTG+GVQA & 60.8 & 72.3 & 50.0 & 59.4 & 66.4 & 42.6 & 38.3 & 36.2 & 16.4 & 70.9 & 55.7 & 60.2 & 38.1 & 66.6 \\
TG+DTG+VHD+GVQA & 60.3 & 71.6 & 48.9 & 60.3 & 67.5 & 44.2 & 37.4 & 35.4 & 16.4 & 66.7 & 70.3 & 76.5 & 60.4 & 68.2 \\
\rowcolor[HTML]{EFEFEF}
TG+DTG+VHD+TAL+GVQA & 61.4 & 72.7 & 51.2 & 59.8 & 66.8 & 43.4 & 39.2 & 35.9 & 17.1 & 70.1 & 71.1 & 79.4 & 61.4 & 71.0 \\ \bottomrule
\end{tabular}}
\caption{
Ablation study on the effect of multi-task reinforcement learning across five temporal understanding tasks. 
Each row indicates the combination of tasks used for training, and the model is evaluated on all tasks. 
Results show that adding more complementary tasks progressively improves overall performance.
}
\label{tab:abtask}
\end{table*}
\subsection{Comparison with State-of-the-arts}

We compare our TempR1 against both expert VLP-based methods and recent MLLM-based methods across five temporal understanding tasks, as reported in Tables~\ref{tab:sota1} and~\ref{tab:sota2}. As shown in Table~\ref{tab:sota1}, TempR1 achieves state-of-the-art performance on Charades-STA~\cite{charadesta} and ActivityNet-Caption~\cite{anetgrounding} for TG, as well as on QVHighlights~\cite{qvhighlights} for VHD. Through multi-task joint training, TempR1 attains 61.4 mIoU on Charades-STA, surpassing VideoChat-R1~\cite{videochat-r1} and MUSEG~\cite{museg} by 0.6 and 1.7 points, respectively. On the more challenging QVHighlights dataset, which involves multi-segment retrieval, TempR1 shows a significantly larger gain, reaching 71.1 mIoU and outperforming the second-best TAR-TVG~\cite{tar-tvg} by 5.2 mIoU. Remarkably, this performance is achieved after training for only one epoch on the combined multi-task dataset. With additional dataset-specific fine-tuning, TempR1 further improves to 62.5 mIoU on Charades-STA (surpassing TAR-TVG by 1.4 points), 51.9 mIoU on ActivityNet, and 71.6 mIoU on QVHighlights. 

As shown in Table~\ref{tab:sota2}, TempR1 also excels on more complex tasks, including DTG, GVQA, and TAL. It achieves 59.8 mIoU on ActivityNet~\cite{anetgrounding} for DTG, comparable to the best VLP expert HSCNet~\cite{hscnet}. On NExTGQA~\cite{nextgqa}, TempR1 obtains 70.1\% question-answering accuracy and 39.2 mIoU for visual evidence localization, improving temporal localization by 3.1 points over VideoChat-R1~\cite{videochat-r1} while maintaining similar answer accuracy. For TAL on ActivityNet-v1.3~\cite{activitynet}, TempR1 achieves 71.0 mF1, outperforming MUSEG~\cite{museg} by a substantial 13.0 points, validating the effectiveness of our prediction-groundtruth matching strategy and reward design.

We further evaluate TempR1 on general video understanding benchmarks, including VideoMME~\cite{videomme}, Perception Test~\cite{perceptiontest}, TempCompass~\cite{tempcompass}, and MVBench~\cite{mvbench}, as illustrated in Figure~\ref{fig:general}. Using the same training data, SFT tends to weaken general reasoning ability compared with the base model, while our reinforcement fine-tuning notably enhances it. This trend aligns with prior observations in VideoChat-R1~\cite{videochat-r1} and Time-R1~\cite{timer1}. Overall, TempR1 delivers the best comprehensive performance across diverse temporal understanding tasks without compromising general video comprehension. These results demonstrate that reinforcement fine-tuning with scalable, task-aligned rewards substantially enhances both temporal comprehension and generalization capabilities of MLLMs in diverse video understanding scenarios.

\subsection{Ablation Study}

\noindent \textbf{Ablation on Localization Reward for TAL.} To evaluate the effectiveness of our Type-3 localization reward design, we conduct an ablation study on ActivityNet-v1.3~\cite{activitynet} for TAL in Table~\ref{tab:abreward}. Removing the instance number reward leads to a noticeable performance drop (70.6 → 60.8 in mF1), confirming that explicitly penalizing instance count mismatches helps the model better estimate the number of actions. Replacing the DP-based matching with a naive sequential matching strategy severely degrades performance to 45.4 mF1, demonstrating the importance of optimal alignment between predicted and ground-truth intervals. Overall, the combination of both components yields the best result, achieving 70.6 mF1 on ActivityNet-v1.3.

\noindent \textbf{Influence of RL Algorithm.}
We evaluate the impact of different RL algorithms on multi-task performance in Table~\ref{tab:abrl}. All three methods achieve strong results across most tasks, highlighting the robustness of multi-task reinforcement learning. Notably, DAPO~\cite{dapo} underperforms on the ActivityNet-v1.3~\cite{activitynet} TAL task in terms of mF1. This is likely due to its token-level policy gradient loss, which biases optimization toward samples with more output instances and affects overall dataset performance. In contrast, GSPO~\cite{gspo} adopts sequence-level optimization, which better balances long and short samples and achieves superior results on multi-segment prediction tasks such as VHD and TAL. To maintain alignment with prior works~\cite{videochat-r1,timer1}, we use GRPO~\cite{deepseekmath} as our default setting.

\noindent \textbf{Effect of Multi-Task Training.} 
Table~\ref{tab:abtask} analyzes the impact of multi-task co-training. As the number of training tasks increases, performance consistently improves across benchmarks, indicating that the model benefits from shared abilities learned from diverse temporal understanding tasks, such as timestamp prediction, text semantic–video segment alignment, and temporal reasoning. Training on all five tasks yields the best results on most metrics, highlighting the synergistic effect of multi-task learning.

\begin{figure*}[t]
       \centering
       \includegraphics[width=0.90\linewidth]{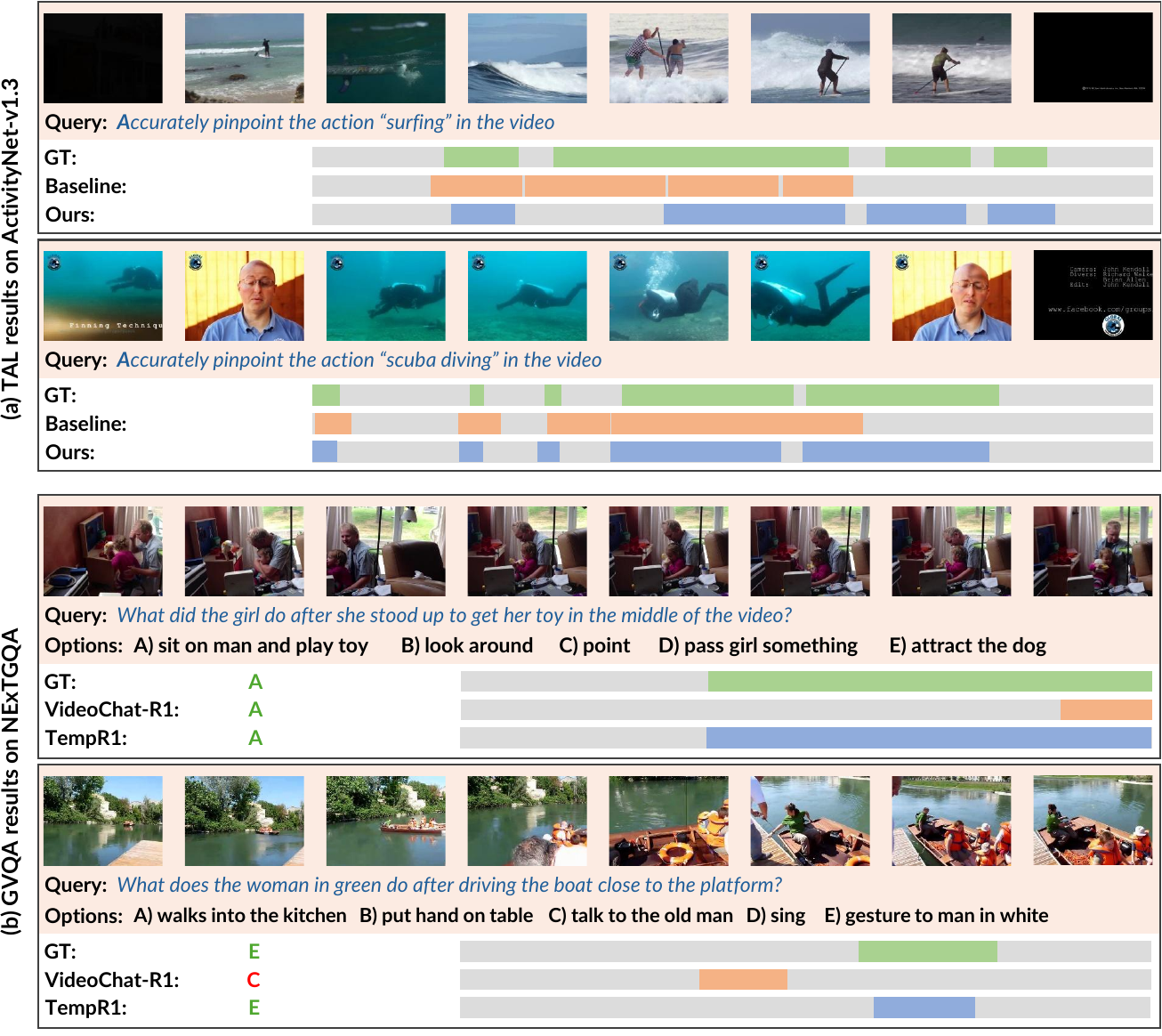}
       \vspace{-5pt}
       \caption{Qualitative result comparisons. (a) Comparison of two matching strategies for localization reward in the TAL task, showing that the DP-based method yields more accurate interval alignment. (b) Comparison of TempR1 and VideoChat-R1 on NExTGQA, where TempR1 provides more precise visual evidence localization.}
       \label{fig:vis}
       \vspace{-5pt}
\end{figure*}

\subsection{Qualitative Results}

We present qualitative comparisons in Figure~\ref{fig:vis} to analyze the effectiveness of the proposed TempR1 model. In Figure~\ref{fig:vis}(a), we compare two matching strategies used for computing the matching reward in the TAL task. The DP-based strategy produces more accurate and consistent temporal localization results in challenging multi-instance scenarios, demonstrating its ability to establish finer correspondences between predicted intervals and ground-truth instances. This leads to more reliable optimization targets during training and improved temporal precision. In Figure~\ref{fig:vis}(b), we compare TempR1 with VideoChat-R1~\cite{videochat-r1} on NExTGQA~\cite{nextgqa}. In the first example, both models predict the correct answer, but TempR1 provides more complete temporal evidence. In the second example, TempR1 shows improved temporal grounding and better reasoning consistency between the visual evidence and textual answer.

\section{Conclusion}

In this paper, we introduce \textbf{TempR1}, a unified multi-task reinforcement learning framework that comprehensively enhances the temporal understanding ability of MLLMs. By jointly training on five diverse tasks with structured, task-aligned rewards under the GRPO framework, TempR1 effectively models diverse temporal structures and achieves consistent performance improvements across benchmarks. Extensive experiments demonstrate that TempR1 outperforms existing SFT- and RL-based approaches, establishing a scalable and effective paradigm for reinforcement fine-tuning in video temporal understanding.

\newpage
{
    \small
    \bibliographystyle{ieeenat_fullname}
    \bibliography{main}
}

\end{document}